\pdfoutput=1
\documentclass[journal]{IEEEtran}
\usepackage{booktabs} 
\usepackage{hyperref}
\usepackage{amsmath}
\usepackage{xcolor}
\hypersetup{
  colorlinks   = true, 
  urlcolor     = blue, 
  linkcolor    = blue, 
  citecolor   = blue 
}
\usepackage{setspace}
\usepackage{graphicx}
\usepackage{array}
\usepackage{multirow}
\usepackage{dblfloatfix}
\usepackage{amssymb}
\ifCLASSINFOpdf
\else
\fi
\begin{document}

\title{Knowledge-enhanced Transformer for Multivariate Long Sequence Time-series Forecasting}

\author{Shubham~Tanaji~Kakde,
        Rony~Mitra,
        Jasashwi~Mandal, and~Manoj~Kumar~Tiwari,~\IEEEmembership{Senior Member,~IEEE}
\thanks{Shubham Tanaji Kakde, Rony Mitra, Jasashwi Mandal, and Manoj Kumar Tiwari are with the Indian Institute of Management Mumbai, Maharashtra, 400087, India. E-mail: kakdeshubham3487@gmail.com; mitrarony92@gmail.com; jasashwi.mandal@gmail.com; mkt09@hotmail.com}}

\maketitle

\begin{abstract}
Multivariate Long Sequence Time-series Forecasting (LSTF) has been a critical task across various real-world applications. Recent advancements focus on the application of transformer architectures attributable to their ability to capture temporal patterns effectively over extended periods. However, these approaches often overlook the inherent relationships and interactions between the input variables that could be drawn from their characteristic properties. In this paper, we aim to bridge this gap by integrating information-rich Knowledge Graph Embeddings (KGE) with state-of-the-art transformer-based architectures. We introduce a novel approach that encapsulates conceptual relationships among variables within a well-defined knowledge graph, forming dynamic and learnable KGEs for seamless integration into the transformer architecture. We investigate the influence of this integration into seminal architectures such as PatchTST, Autoformer, Informer, and Vanilla Transformer. Furthermore, we thoroughly investigate the performance of these knowledge-enhanced architectures along with their original implementations for long forecasting horizons and demonstrate significant improvement in the benchmark results. This enhancement empowers transformer-based architectures to address the inherent structural relation between variables. Our knowledge-enhanced approach improves the accuracy of multivariate LSTF by capturing complex temporal and relational dynamics across multiple domains. To substantiate the validity of our model, we conduct comprehensive experiments using Weather and Electric Transformer Temperature (ETT) datasets.

\end{abstract}

\begin{IEEEkeywords}
Transformer, Knowledge Graphs, Time-series Forecasting
\end{IEEEkeywords}

\IEEEpeerreviewmaketitle

\section{Introduction}
\IEEEPARstart{R}{esearch} in multivariate Long Sequence Time-series Forecasting (LSTF) has gained prominence in recent years due to its wide practical applications \cite{mlp_forecasting}, \cite{tsf_review}, \cite{tsf_dl_review}. Real-world applications of time-series forecasting span a wide range of domains, including weather and climate forecasting \cite{weather_research2}, energy grid analysis \cite{smart_grid}, \cite{energyload2}, traffic volume estimation \cite{trafficflow1}, \cite{trafficflow2}, \cite{trafficflow3}, retail sales forecasting \cite{retail1} and financial market predictions \cite{financial1}, \cite{financial2}. These comprehensive applications highlight the continuing significance of time-series forecasting and provide scope for evolving methodologies to address these applications. Furthermore, the time-series forecasting problem itself presents significant challenges due to its intricate temporal patterns and complex dependencies among variables. Incorporating the additional parameter of multivariable long-sequence forecasting adds a layer of complexity to the analysis. Numerous classical methods, including statistical models such as Vector Autoregression (VAR) \cite{var}, Gaussian Process \cite{gp}, Support Vector Regression \cite{svr} and deep learned-based models such as Multi-layer perceptrons (MLP), Recurrent Neural Networks (RNNs), Long-Short Term Memory (LSTM) and Convolutional Neural Networks (CNNs) have been employed extensively and have demonstrated significant potential in short sequence forecasting across various domains.  
    
However, classical methods are not devoid of limitations. They often struggle to capture complex temporal patterns. Additionally, the non-stationarity of real-world time-series data poses a significant challenge thereby resulting in lack of robustness. Thus, accurately quantifying complex multivariate relationships continues to pose formidable challenge. For deep learning techniques in the context of LSTF, issues such as suboptimal parameter estimation, poor generalization and gradient instability are frequently encountered.
\begin{figure}
\centering
\includegraphics[width=1\linewidth]{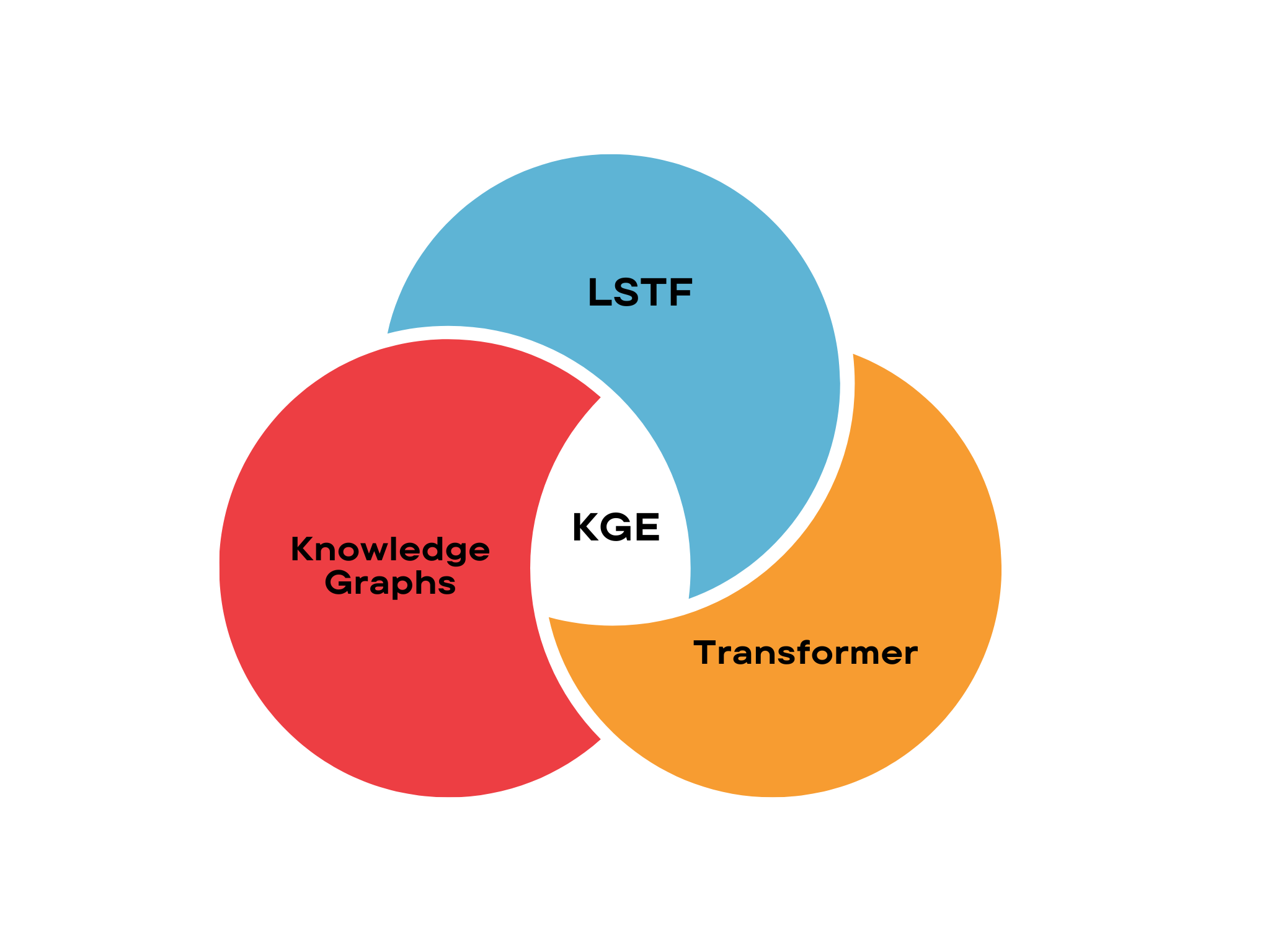}
    \caption{Venn diagram illustrating the central role of Knowledge Graph Embeddings at the intersection of Knowledge Graphs, Long Sequence Time-series Forecasting, and Transformer architectures}
    \label{fig:1}
\end{figure}

With the advent of enhanced parallel computational systems \cite{nvidia} and the groundbreaking success of the Transformer \cite{transformer} architecture, the landscape of research in time-series forecasting has undergone a significant transformation. These attention mechanism-based architectures have demonstrated an exceptional ability to discern critical patterns in sequential data, thus marking a pivotal shift in the field. Studies such as Informer \cite{informer}, Autoformer \cite{autoformer}, FEDFormer \cite{fedformer} and PatchTST \cite{patchtst} have been seminal applications of the Transformer architecture for long sequence time-series forecasting (LSTF). These studies have surpassed the conventional forecasting methods on various benchmark datasets from real-world applications such as \textit{Weather} \cite{weather_data}, \textit{Traffic} \cite{traffic_data}, \textit{ETT} \cite{informer}, \textit{Electricity} \cite{electricity_data}, \textit{ILI} \cite{ILI_data} and many more. The attention mechanisms are adept at analyzing multivariate time-series data and are highly effective in identifying temporal patterns on an individual timestamp level. However, it is pertinent to recognize that the attention mechanism can overlook crucial multivariate structural information, particularly in long-term forecasting \cite{att_failure_tsf}. This oversight is especially relevant to the nuanced structural or spatial relationships crucial for a comprehensive understanding of the dataset \cite{relphormer}. To address this issue, incorporating spatio-temporal training in LSTF problem offers a compelling opportunity to explore the integration of spatial relations alongside temporal dependencies for more accurate forecasting.

Graph-based methodologies in LSTF have attracted considerable interest owing to their potential in modeling complex interactions between spatial and temporal dependencies \cite{spacetimetransformer}, \cite{DKG_mLSTF}, \cite{DKG_mLSTF1}, \cite{kde_gan_lstf}. However, current research heavily relies on dynamic knowledge graph frameworks and computationally intensive Graph Neural Networks (GNN) or intricate Graph Convolutional Networks (GCN) \cite{kde_dynamic_graph}, \cite{kde_gnn}. Despite their complexity, these approaches frequently fall short when assessed against standardized configurations of state-of-the-art (SOTA) models like PatchTST, Autoformer, Informer and FedFormer, particularly for longer forecast horizons. Researchers have primarily emphasized dynamic knowledge graphs to comprehend spatio-temporal dynamics. The integration of intuitive and conventional knowledge graphs is less evident due to the superior performance of dynamic counterparts in temporal tasks. Hence, there is a significant gap in the research landscape concerning experiments with the integration of knowledge graphs with complex deep-learning architectures for temporal tasks.

Our proposed approach aims to overcome the constraints of current graph-based architectures in LSTF by introducing a novel approach that encapsulates inherent relationships among variables within a knowledge graph and integrates them into transformer-based frameworks by forming dynamic and learnable Knowledge Graph Embeddings (KGEs). In the context of multivariate LSTF, we construct a basis for initial information relationships among the variables in the form of a knowledge graph by deriving conceptual relationships from literature, further producing the dynamic KGE. These embeddings augment transformer-based architectures with crucial spatial relations, thereby enhancing forecasting accuracy without introducing unnecessary complexity. The construction of the knowledge graph entails a comprehensive analysis of the multivariate time series to ensure its efficacy in capturing relevant domain knowledge. The main contributions of the papers are highlighted as follows:
\begin{enumerate}
    \item This paper introduces an intuitive knowledge-graph enhancement within SOTA transformer-based architectures for effective forecasting on multivariate Long Sequence Timeseries Forecasting (LSTF) problems.

    \item We propose a novel approach that constructs the conceptual relations of variables in the form of a well-defined knowledge graph and develops dynamic and learnable knowledge graph embeddings, enhancing the spatio-temporal relationships in SOTA transformer architectures.
 
    \item Through rigorous experimentation on longer forecast horizons, this study unveils the substantial potential of this approach, paving the way for further advancements and refinement in time-series forecasting methodologies.
\end{enumerate}

To the best of our knowledge, this study is the first attempt to integrate dynamic knowledge graph embeddings to transformers based on the initial information from a knowledge graph for LSTF problem. It is also pertinent to note that the primary objective of the study is to refine the accuracy of LTSF by introducing a novel learnable knowledge graph embedding approach in the baseline architectures rather than focusing on uncovering the ground-truth graph structure.

In the following sections, we conduct a thorough exploration of the existing literature in section \ref{sec:literature_review}, followed by a detailed exposition of our methodology in section \ref{sec:methodology}. We then present a comprehensive analysis of our experimental findings in section \ref{sec:experiments}. Finally, we conclude by reflecting on the implications of our study's results and outline the potential directions for future research in the concluding section \ref{sec:conclusion}. 

\section{Related Work}
\label{sec:literature_review}

Time-series forecasting has been a pivotal research area in predictive analytics for decades, owing to its ubiquitous relevance across fields like finance, economics, meteorology and manufacturing. Early researchers have pioneered canonical statistical methodologies, including Autoregressive Integrated Moving Average (ARIMA) models \cite{boxj}, Exponential Smoothing techniques \cite{expsmooth}, Seasonal Decomposition of Time Series (STL) \cite{STL}, Gaussian Processes (GP) \cite{gp}, Bayesian Structural (BSTS) Models \cite{BSTS} and Dynamic Linear Models (DLM) \cite{DLM}. These traditional techniques have long dominated the realm of time-series analysis owing to their interpretability, simplicity and historical performance.

However, the landscape of data has evolved dramatically, marked by unprecedented volumes, varieties and velocities. Traditional statistical methods excel at capturing linear patterns and simple seasonal variations, yet they encounter difficulties accommodating the inherent complexities of time-series data. These approaches often struggle to fully mitigate challenges stemming from irregular fluctuations, non-stationarity, high-frequency data and intricate interdependencies. To address these issues, researchers and practitioners have gravitated towards exploiting deep learning techniques like Convolutional Neural Networks (CNNs), Recurrent Neural Networks (RNNs), Long Short-Term Memory (LSTM) networks and hybrid approaches that blend statistical methods with machine learning algorithms. These methodologies capture the nuanced patterns present in dynamic time-series data, thus offering greater efficacy. 

With the emergence of robust parallel computing systems \cite{nvidia} and the notable advancements in the Transformer \cite{transformer} architecture, the field of time-series forecasting underwent a significant evolution. Numerous studies \cite{informer}, \cite{autoformer}, \cite{fedformer}, \cite{patchtst}, \cite{pyraformer} capitalized on the canonical attention mechanism of transformer for timeseries forecasting problems and demonstrated superior performance compared to both traditional and other deep learning approaches. However, particularly remarkable is their efficacy in addressing the intricacies of Long Sequence Timeseries Forecasting (LSTF), a task where conventional techniques frequently encounter difficulties. The seminal work of LSTNet \cite{LSTNET} revolutionized the understanding of short-term and long-term temporal patterns by introducing a temporal attention layer. This innovative integration amplified the efficacy of deep learning features extracted from CNN and RNN, enabling the adept capture of both short-term local dependencies and long-term trends within multivariate time-series data.

In the realm of transformer architectures within LSTF, a surge of research has concentrated on refining attention mechanisms to streamline training and reduce time complexities. Informer \cite{informer} introduced an enhanced \textit{ProbSparse} self-attention mechanism to mitigate the quadratic time complexity and high memory consumption challenges inherent in Transformer. Similarly, Pyraformer \cite{pyraformer} proposed a pyramidal attention module for effectively capturing both short and long-term temporal dependencies in timeseries data with reduced space-time complexities. Alternative approaches for attention mechanisms such as Local-sensitive hashing \cite{reformer} and Attention-free LSTM blocks \cite{attentionfree} have demonstrated competitive performances. Beyond the modifications of the attention mechanisms, significant studies \cite{improvedposencoding}, \cite{cape}, \cite{roformer} have focused on enhancing the novel positional encoding of the transformer to elevate performance results on time-series forecasting and Neural Machine Translation (NMT) tasks.  

In recent years, there has been a notable trend towards integrating statistical decomposition methods with attention blocks to tackle non-stationarity in timeseries data. This convergence of attention mechanisms with statistical concepts like non-stationarity and seasonality-trend decomposition techniques marks a promising advancement in the field of time-series forecasting methodologies. Autoformer \cite{autoformer} addressed the LSTF problem through a decomposition architecture, drawing inspiration from established decomposition approaches and renovating the self-attention mechanism by incorporating auto-correlation blocks. Similarly, FEDformer \cite{fedformer} proposed a mixture-of-experts framework in the transformer to enhance seasonal-trend decomposition components using Fourier and Wavelet blocks, which could efficiently capture global trends in time-series data. Additionally, Fedformer achieves a linear computational complexity, ensuring both effectiveness and efficiency. Non-stationary Transformer \cite{nonstationarytransformer} addressed the challenges of non-stationarity in real-world data by introducing a de-stationary attention module, effectively preserving intrinsic non-stationarity while enhancing predictive accuracy. PatchTST \cite{patchtst} introduced a novel approach to time-series forecasting by segmenting data into subseries-level patches and ensuring channel independence. This enables the model to efficiently capture the local semantic information in order to attend to a longer historical context. This design significantly reduces computational complexity and outperforms previous architectures on large datasets. 

The advancements in transformer-based architectures for forecasting present a strong case for their robustness and reliability. However, while attention mechanisms excel at capturing temporal patterns, they often overlook essential multivariate structural information. To bridge this gap, researchers are actively exploring spatio-temporal architectures within LSTF. Notably, the application of graph-based methodologies like Graph Neural Networks (GNN) \cite{GNN}, Graph Convolutional Networks \cite{GCN} have emerged as promising avenues. These methodologies offer the potential to seamlessly integrate spatial and temporal dynamics, enhancing the predictive capabilities of forecasting models. The utilization of graph-based methodologies within sequential deep learning architectures for timeseries forecasting has garnered considerable attention in the literature. GNN and GCN frameworks are frequently employed to facilitate multivariate timeseries forecasting by autonomously discerning spatial and temporal interdependencies among variables \cite{kde_gnn}, \cite{kde_gan}, \cite{dyn_graph_nn}, \cite{dynamic_hypergraph}, \cite{DSTAGCN}, \cite{stgcn} \cite{stemgnn}. 

Notably, DHSL \cite{dynamic_hypergraph} model captured high-order correlations in multivariate time series forecasting by generating and optimizing dynamic hypergraph structures using the K-Nearest Neighbors method. DSTAGCN \cite{DSTAGCN} highlights dynamic spatial-temporal dependencies in traffic forecasting by linking the latest time slice to the past by leveraging a fuzzy neural network to generate a dynamic adjacency matrix for enhanced performance. STGCN \cite{stgcn} devised a comprehensive convolution-based structure for their forecasting model, resulting in expedited training owing to fewer parameters while maintaining competitive forecasting outcomes. StemGNN \cite{stemgnn} introduced a spectral temporal GNN that amalgamates the Graph Fourier Transform and Discrete Fourier Transform to capture cross-correlations and temporal dependencies in multivariate timeseries data. Moreover, the integration of learnable graph blocks within a transformer framework has emerged as a burgeoning research domain \cite{spacetimetransformer}, \cite{kde_gan_lstf}, \cite{kde_gan}, \cite{attention_gcn}, \cite{graphtransformer}, \cite{graphformer}. Particularly, Spacetimeformer \cite{spacetimetransformer} and Graphformer \cite{graphformer} advocate for this approach, outlining the incorporation of graph convolution blocks in a transformer setting for forecasting tasks. These methodologies typically harness dynamic interrelationships among variables acquired through the GNN or GCN blocks.

However, while endeavoring to capture the intricate spatio-temporal dynamics of timeseries data, they often manifest over-generalization, thereby diminishing forecast accuracies. Furthermore, these approaches perform well in short-term forecasting but often exhibit notable errors for longer forecast horizons \cite{stgcn}. Comparative evaluations against equivalently sized graph-based transformer within the same setting highlight the superior performance of conventional transformer-based on the considered benchmark datasets \cite{autoformer}, \cite{patchtst}, \cite{spacetimetransformer}, \cite{kde_gan}, \cite{graphformer}. 

Besides, it is noteworthy that prior studies have primarily focused on automatically constructing dynamic graphs from the data. However, there remains a considerable dearth of analysis concerning the variables within a multivariate timeseries and deriving inherent relationships solely from those variables. In this study, we propose leveraging learnable knowledge graph embeddings to effectively capture the inherent spatial relationships among variables while harnessing temporal dependencies from SOTA transformer-based architectures.

\begin{figure*}
    \centering
    \includegraphics[width=0.81\linewidth]{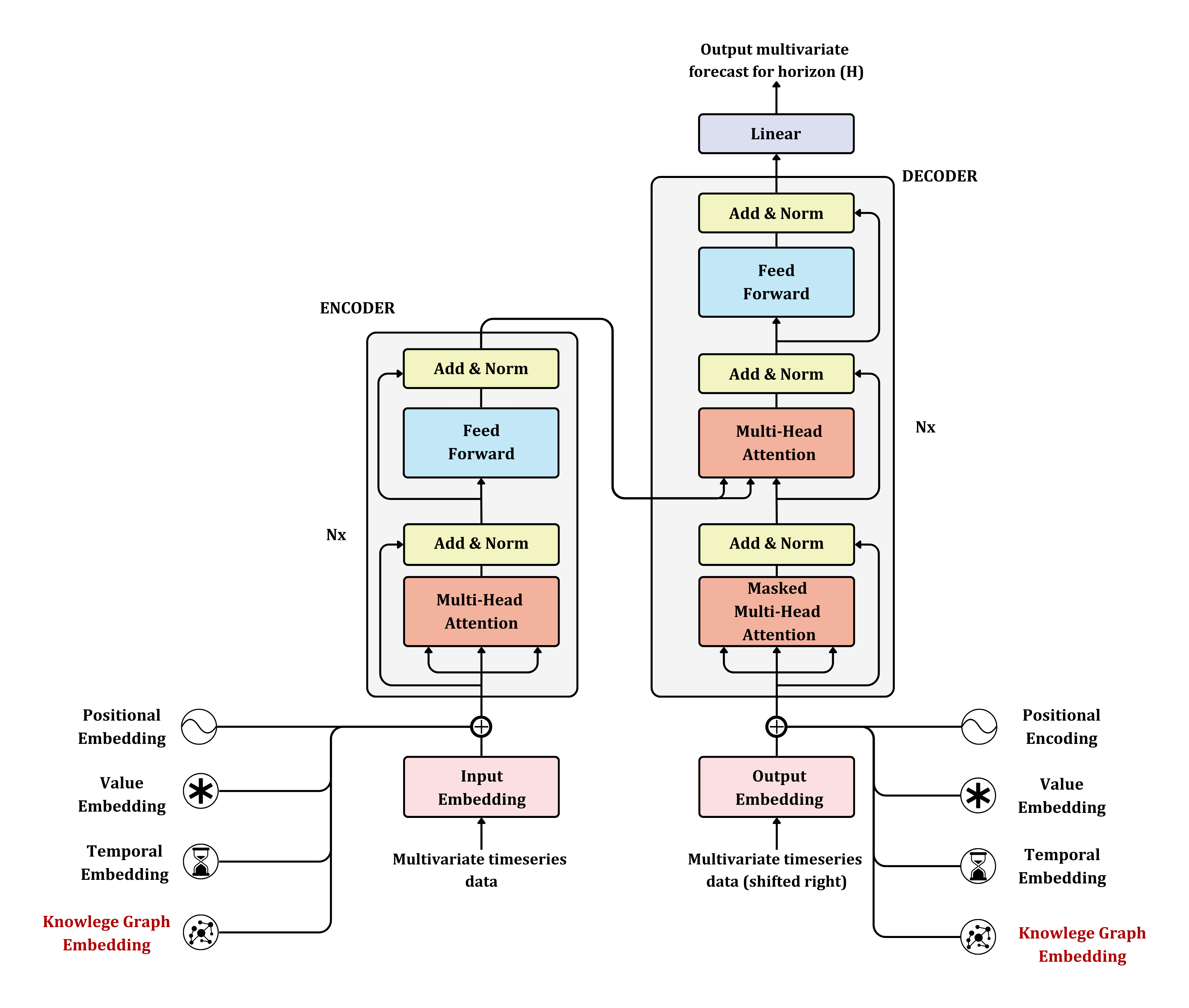}
    \caption{Model structure of knowledge-enhanced transformer for LSTF; Knowledge graph embeddings along with Positional embeddings, Value embeddings and Temporal embedding are added with input embeddings towards encoder and decoder blocks}
    \label{fig:2}
\end{figure*}

\section{Proposed Methodology}
\label{sec:methodology}
We consider the following problem: For a set of multivariate time-series dataset $\mathbf{X}=\left\{ \mathbf{X}^{(1)}_t, \ldots, \mathbf{X}^{(M)}_t \right\}_{t=1}^L$, where \(M\) is the number of variates, \(L\) is the look-back window and $\mathbf{X}^{(i)}_t$ denotes the $i^{\text{th}}$ variable in $t^{\text{th}}$ timestamp.  The Long Sequence Time-series Forecasting (LSTF) task is to forecast the set $\hat{\mathbf{X}} = \left\{ \hat{\mathbf{X}}^{(1)}_{t}, \ldots, \hat{\mathbf{X}}^{(M)}_{t} \right\}_{t=L+1}^{L+H}$, wherein \(H\) denotes the prediction horizon. In this study, we address the Direct Multi-step Estimation \cite{dms} approach, which entails considering the entire forecasting horizon simultaneously within the objective function. This approach is implemented to enhance computational efficiency and minimize the risk of bias accumulation, as observed in Iterative Multi-Step (IMS) forecasting \cite{ims_vs_dms1}, \cite{dlinear}.

{\subsection{Model Structure}}
In this multivariate LSTF problem, we propose an integration of learnable knowledge graph embeddings with novel architectures (\textit{Transformer, Autoformer, Informer, PatchTST}). For illustration purposes, we exemplify the integration of knowledge graph embeddings (KGE) with the original Transformer \cite{transformer} architecture. A detailed flow of the model structure is shown in Fig. \ref{fig:2}.  

\subsubsection{Input Embeddings}
The input embeddings are enhanced by the proposed knowledge graph embedding with the existing embeddings as given in equations \ref{eqn1} and \ref{eqn2} :
\begin{equation}
    \mathbf{W}_{emb} = \mathbf{W}_{KGE}+\mathbf{W}_{PE}+\mathbf{W}_{VE}+\mathbf{W}_{TE}
    \label{eqn1}
\end{equation}
\begin{equation}
    \mathbf{Z} = \mathbf{X} + \mathbf{W}_{emb}
    \label{eqn2}
\end{equation}
Here, 
\begin{itemize}
    \item $\mathbf{W}_{KGE}$ refers to the proposed knowledge graph embedding. It represents the inherent domain's structural knowledge in the variables
    \item $\mathbf{W}_{PE}$ is the sinusoidal positional embedding \cite{transformer} which captures the order of the sequence
    \item $\mathbf{W}_{VE}$ is convoluted value embeddings \cite{informer} used for representing the magnitudes of the time-series variables
    \item $\mathbf{W}_{TE}$ \cite{dlinear} refers to temporal embeddings constructed from timestamps to encode temporal patterns
    \item $\mathbf{Z}$ is the enhanced input sequence that incorporates element-wise aggregation between $\mathbf{X}$ and $\mathbf{W}_{emb}$
\end{itemize}

These embeddings are selectively employed by the existing benchmark architectures. In our replication of the benchmark results, we adhere to this judicious approach to embedding integration with KGE.

\begin{figure*}[h]
    \centering
    \includegraphics[width=0.8\linewidth]{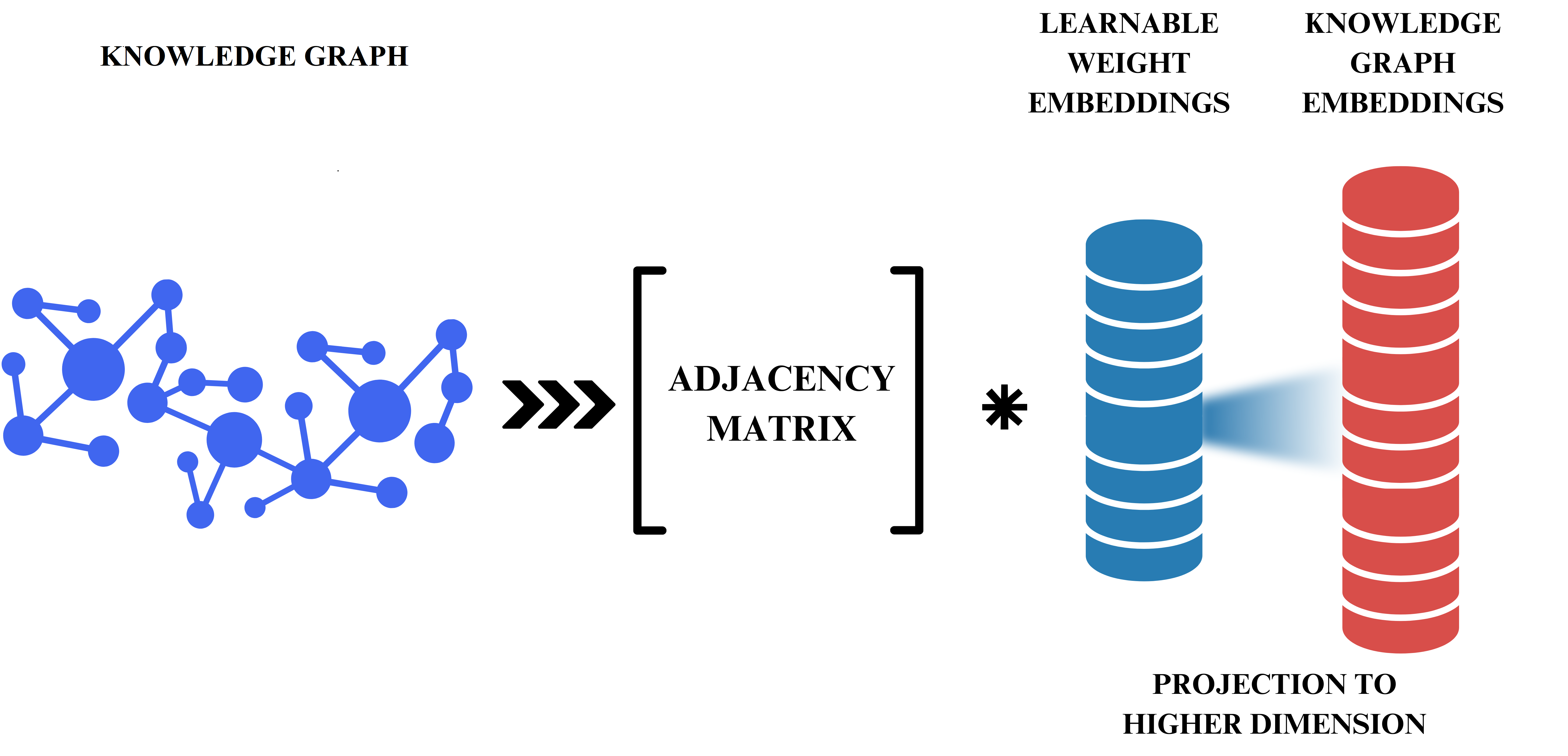}
    \caption{Flowchart for construction of learnable Knowledge Graph Embeddings (KGE)}
    \label{fig:3}
\end{figure*}

\subsubsection{Multi-Head Attention in Encoder-Decoder}

The encoder-decoder blocks are the core of the Transformer-based architecture. We employ the canonical multi-head attention (MHA) mechanism \cite{transformer} in these encoder-decoder blocks. For this, we first perform the linear projections on input embeddings from equation \ref{eqn2} to obtain the following matrices:
\begin{equation}
    \mathbf{Q} = \mathbf{Z}\mathbf{W}^{Q};\quad \mathbf{K} = \mathbf{Z}\mathbf{W}^{K};\quad\mathbf{V} = \mathbf{Z}\mathbf{W}^{V} 
\label{eqn3}
\end{equation}
Here, $\mathbf{Q}$, $\mathbf{K}$, $\mathbf{V}$ are the query, key and value matrices derived from the input embeddings such that $\mathbf{W}^{Q} \in \mathbb{R}^{D \times d_k}, \quad \mathbf{W}^{K} \in \mathbb{R}^{D \times d_k}, \quad \mathbf{W}^{V} \in \mathbb{R}^{D \times d_v}$

\begin{itemize}
    \item \(d_k\) is the dimensionality of the query and key matrices
    \item \(d_v\) is the dimensionality of the value matrix
    \item \(D\) is the dimensionality of the model
\end{itemize}

The attention score for each head is calculated through the weighted aggregation of a scaled dot-product of the query and key matrices, with the value matrices acting as weights. This is followed by a softmax operation to normalize the scores, given in equation \ref{eqn5}. 
\begin{equation}
    \text{head}_i=\text{Attention}(\mathbf{Q}_{i}, \mathbf{K}_{i}, \mathbf{V}_{i})
    \label{eqn4}
\end{equation}
\begin{equation}
\text{Attention}(\mathbf{Q}_{i}, \mathbf{K}_{i}, \mathbf{V}_{i})=\text{softmax}\left(\frac{\mathbf{Q}_{i}\mathbf{K}_{i}^T}{\sqrt{d_k}}\right)\mathbf{V}_{i} \label{eqn5}
\end{equation}
\begin{equation}
    \text{Multihead}(\mathbf{Q}, \mathbf{K}, \mathbf{V})=\text{Concat}(\text{head}_{1}, \ldots, \text{head}_{h})\mathbf{W}_{O}
    \label{eqn6}
\end{equation}
In equation \ref{eqn6}, $\mathbf{W}_{O}\in\mathbb{R}^{h d_v \times D}$, where \(h\) is the number of heads in multi-head attention. This segmentation of the single attention head into multiple heads facilitates the simultaneous attention to information from diverse representational subspaces. As this employs a straightforward dimensionality reduction, it does not burden the total computational cost \cite{transformer}. The encoder-decoder components are highlighted in the knowledge-enhanced transformer in Fig. \ref{fig:2}. Here, MHA allows each position in the encoder-decoder to attend to all positions from the input sequence. Moreover, in the decoder, the masking of future timestamps is enabled to avoid attending to unseen information. This is crucial for ensuring causality and maintaining the autoregressive property of the model, where each prediction can only depend on past and present information.

 Similar to Transformer, we incorporate KGE with Autoformer, Informer and PatchTST and provide a comparative study in further sections.
 
\subsubsection{Loss Function}
Mean Squared Error (MSE) loss is used to assess the deviation between the prediction and the ground truth. The loss across each channel is aggregated and then averaged across M time-series to obtain the overall objective loss, given by equation \ref{eqn7}.

\begin{equation}
\mathcal{L} = \frac{1}{M} \sum_{i=1}^{M} \left\| \hat{\mathbf{X}}^{(i)}_{L+1:L+H} - \mathbf{X}^{(i)}_{L+1:L+H} \right\|^2
    \label{eqn7}
\end{equation}
where $\mathbf{\hat{X}}^{(i)}_{L+1:L+H}$ represent the prediction and $\mathbf{X}^{(i)}_{L+1:L+H}$ denotes the actual ground truth values.

{\subsection{Construction of Dynamic and Learnable Knowledge Graph Embeddings from Knowledge Graph}}
To leverage the full potential of conceptual and spatio-temporal relations among variables, we propose a hybrid approach that highlights conceptual knowledge graphs to form information-rich dynamic knowledge graph embeddings. The dynamic influence in the KGEs is induced by the backpropagation process in the transformer-based architectures. We define our knowledge graph $\mathbf{G}=(\mathbf{V},\mathbf{E})$ for the multivariate time-series such that $\mathbf{V}$ represents a set of nodes each representing a variable in $\mathbf{X}$. 
In other words, $\mathbf{V}=\left\{ \mathbf{v}^{(1)}, \ldots, \mathbf{v}^{(M)} \right\}$ where node $\mathbf{v}^{(i)}$ corresponds to $\mathbf{X}^{(i)}_t$ and $\mathbf{e}_{ij}$  $\in$  $\mathbf{E}$ where $\mathbf{e}_{ij}$ represent edge between nodes  $\mathbf{v}^{(i)}$ and $\mathbf{v}^{(j)}$. In this knowledge graph, we construct an adjacency matrix $\mathbf{A}$ as a binary matrix representing the presence of edges between nodes. Thus, $\mathbf{A} = \{\mathbf{a}_{ij}\} \in \{0, 1\}^{|\mathbf{V}| \times |\mathbf{V}|}$, where each element $\mathbf{a}_{ij}$ is set to 1 if there is an edge between nodes $\mathbf{v}^{(i)}$ and $\mathbf{v}^{(j)}$ and 0 otherwise. 

To construct dynamic and learnable knowledge graph embeddings $\mathbf{W}_{KGE}$, we employ a sparse matrix transformation to convert the adjacency matrix into learnable weights. The adjacency matrix $\mathbf{A}$ is multiplied by a learnable weight matrix $\mathbf{W}_l$, followed by an Einstein summation with a learnable projection weight matrix $\mathbf{W}_p$. This projection matrix is introduced to align with the shape of the positional embeddings. Here, $\mathbf{W}_l\in\mathbb{R}^{V \times D}$ and $\mathbf{W}_p\in\mathbb{R}^{L \times D}$. Fig. \ref{fig:3} shows the visualization for the construction of $\mathbf{W}_{KGE}$. 
\\
\section{Experiments}
\label{sec:experiments}
\subsection{Datasets}

The proposed knowledge graph embedding strategy is evaluated on five publicly available datasets, including Weather \cite{weather_data} and 4 ETT (Electric Transformer Temperature) \cite{informer} datasets (ETTm1, ETTm2, ETTh1, ETTh2). These datasets are widely accepted as standard benchmarks for testing the performance of novel architectures \cite{informer}, \cite{autoformer}, \cite{fedformer}, \cite{patchtst}, \cite{dlinear}. Table \ref{tab1} depicts the statistical highlights of the benchmark datasets. Our evaluation encompasses a diverse range of dataset sizes to ensure a comprehensive assessment of our proposed integration. Moreover, we have excluded datasets that either contain limited data or lack conceptual or any existing structural relationships between their variables, as these characteristics are crucial for the effective application of our methodology \cite{traffic_data}, \cite{electricity_data}, \cite{ILI_data}.

\begin{table}[h]
\caption{Statistical highlights of benchmark datasets}
\centering
\begin{tabular}{l c c c}
\toprule
\textbf{Datasets} & \textbf{Parameters} & \textbf{Frequency} & \textbf{Timestamps} \\
\midrule
Weather & 21 & 10 mins & 52696 \\
ETTm1 & 7 & 15 mins & 69680 \\
ETTm2 & 7 & 15 mins & 69680 \\
ETTh1 & 7 & 1 hour & 17420 \\ 
ETTh2 & 7 & 1 hour & 17420 \\
\bottomrule
\end{tabular}
\label{tab1}
\end{table}


\subsubsection{Weather} The Weather dataset is a freely-accessible repository recorded, preprocessed and documented by Max Planck Institute for Biogeochemistry, Germany \cite{weather_data}. A total of 21 weather indicator readings have been updated regularly every 10 minutes since 2008. We consider the dataset for the year 2020 as adopted by benchmark architectures for the LSTF study. Table \ref{tab2} depicts the indicators taken into consideration in this study. 

\subsubsection{ETT}
The Electric Transformer Temperature (ETT) dataset, as documented in the literature \cite{informer}, encompasses data collected from two distinct counties in China over the period from 2016 to 2018. This dataset captures the Oil Temperature (OT) alongside six features related to power load, namely - HUFL (High Useful Load), HULL (High Useless Load), MUFL (Middle Useful Load), MULL (Middle Useless Load), LUFL (Low Useful Load) and LULL (Low Useless Load). It is organized into two subsets: ETTh and ETTm, which denote the temporal resolution of the data in hours (ETTh) and minutes (ETTm), respectively.

\begin{table}[h]
\centering
\caption{Weather Dataset Indicators}
\renewcommand{\arraystretch}{1.15}
\begin{tabular}{p{1.5cm}p{4.3cm}p{2cm}}
\toprule
Symbol & Description & Unit \\
\midrule
\( Timestamp \) & Timestamp of the data record & DD.MM.YYYY HH:MM \\
\( p \) & Air pressure & mbar \\
\( V_{Pdef} \) & Water vapor pressure (Deficit) & mbar\\
\( V_{Pact} \) & Water vapor pressure (Actual) & mbar\\
\( V_{Pmax} \) & Water vapor pressure (Saturation) & mbar\\
\( T \) & Air temperature & \( ^\circ \)C\\
\( T_{dew} \) & Dew point temperature & \( ^\circ \)C\\
\( T_{pot} \) & Potential temperature & K\\
\( T_{log} \) & Internal logger temperature & \( ^\circ \)C\\
\( sh \) & Specific humidity & g kg\(^{-1}\)\\
\( rh \) & Relative humidity & \%\\
\( rho \) & Air density & g m\(^{-3}\)\\
\( H_{2}OC \) & Water vapor concentration & mmol mol\(^{-1}\)\\
\( rain \) & Precipitation  & mm\\
\( raining \) & Duration of precipitation & s\\
\( PAR \) & Photosynthetically Active Radiation & \( \mu \)molm\(^{-2}\)s\(^{-1}\) \\
\( SWDR \) & Short Wave Downward Radiation & W m\(^{-2}\)\\
\( max. PAR \) & Photosynthetically Active Radiation (Max) & \( \mu \)molm\(^{-2}\) s\(^{-1}\)\\
\( wd \) & Wind direction & \( ^\circ \)\\
\( wv \) & Wind velocity (Actual) & m s\(^{-1}\)\\
\( max. wv \) & Wind velocity (Max) & m s\(^{-1}\)\\
\( CO_2 \) & \( CO_2 \)-Concentration of ambient air & ppm\\
\bottomrule
\end{tabular}
\label{tab2}
\end{table}

\subsection{Knowledge Graphs for benchmark datasets}

In this study, we utilize knowledge graph embeddings (KGEs), initially derived from well-defined conceptual knowledge graphs and projected to higher dimensions for infusing the dynamic nature tailored fit to individual datasets. Unlike conventional approaches that employ temporal graphs or graph convolution techniques to capture spatio-temporal dynamics, our method leverages a simple and intuitive knowledge graph for spatial features and integrates it with the novel attention-based mechanism in the form of dynamic KGEs to identify temporal patterns. This approach facilitates the generation of the dynamic embeddings through the extraction of factual relationships among the variables.

\begin{table*}[!t]
\centering
\caption{Performance Comparison of Models with and without Knowledge Graph Embeddings (Transformer, Informer and Autoformer)}
\label{tab3.1}
\resizebox{\linewidth}{!}{%
\begin{tabular}{|c|c||c|c|c|c||c|c|c|c||c|c|c|c||}
\hline
\multirow{1}{*}{Dataset} & \multirow{1}{*}{Horizon}& \multicolumn{2}{c|}{\shortstack[c]{\\[0.1ex]Transformer\\(With KGE)\\[0.1ex]}} & \multicolumn{2}{c||}{\shortstack[c]{\\[0.1ex]Transformer\\(Original)\\[0.1ex]}
} & \multicolumn{2}{c|}{\shortstack[c]{\\[0.1ex]Informer\\(With KGE)\\[0.1ex]}} & \multicolumn{2}{c||}{\shortstack[c]{\\[0.1ex]Informer\\(Original)\\[0.1ex]}
} & \multicolumn{2}{c|}{\shortstack[c]{\\[0.1ex]Autoformer\\(With KGE)\\[0.1ex]}} & \multicolumn{2}{c||}{\shortstack[c]{\\[0.1ex]Autoformer\\(Original)\\[0.1ex]}
} \\ \cline{3-14} 
                          &                           & \multicolumn{1}{l|}{MSE} & \multicolumn{1}{l|}{MAE} & \multicolumn{1}{l|}{MSE} & \multicolumn{1}{l||}{MAE
} & \multicolumn{1}{l|}{MSE} & \multicolumn{1}{l|}{MAE} & \multicolumn{1}{l|}{MSE} & \multicolumn{1}{l||}{MAE
} & \multicolumn{1}{l|}{MSE} & \multicolumn{1}{l|}{MAE} & \multicolumn{1}{l|}{MSE} & \multicolumn{1}{l||}{MAE
}    \\ \hline

\multirow{4}{*}{Weather} & 96 & \textbf{0.264} & \textbf{0.337} & 0.400 & 0.450 
& \textbf{0.198} & \textbf{0.275} & 0.316 & 0.367 
& \textbf{0.261} & \textbf{0.323} & 0.289 & 0.357 
\\
 & 192 & \textbf{0.310} & \textbf{0.368} & 0.364 & 0.437 
& \textbf{0.254} & \textbf{0.317} & 0.467 & 0.471 
& \textbf{0.288} & \textbf{0.335} & 0.312 & 0.364 
\\
 & 336 & \textbf{0.360} & \textbf{0.395} & 0.448 & 0.495 
& \textbf{0.299} & \textbf{0.348} & 0.788 & 0.644 
& \textbf{0.320} & \textbf{0.353} & 0.349 & 0.387 
\\
 & 720 & \textbf{0.434} & \textbf{0.458} & 0.555 & 0.547 
& \textbf{0.353} & \textbf{0.399} & 1.301 & 0.860 
& \textbf{0.371} & \textbf{0.387} & 0.393 & 0.412 
\\ \hline
\multirow{4}{*}{ETTm1} & 96 & 0.863 & 0.681 & \textbf{0.831} & \textbf{0.680} 
& \textbf{0.733} & \textbf{0.615} & 0.839 & 0.674 
& \textbf{0.453} & \textbf{0.460} & 0.664 & 0.539 
\\
 & 192 & \textbf{0.888} & \textbf{0.698} & 0.982 & 0.773 
& \textbf{0.806} & \textbf{0.682} & 0.903 & 0.734 
& \textbf{0.467} & \textbf{0.462} & 0.682 & 0.547 
\\
 & 336 & \textbf{0.974} & \textbf{0.715} & 1.079 & 0.846 
& \textbf{0.947} & \textbf{0.756} & 1.220 & 0.877 
& \textbf{0.631} & \textbf{0.533} & 0.686 & 0.565 
\\
 & 720 & 1.030 & 0.769 & \textbf{0.975} & \textbf{0.750} 
& 1.461 & 0.949 & \textbf{1.031} & \textbf{0.762} 
& \textbf{0.636} & \textbf{0.539} & 0.759 & 0.597 
\\ \hline
\multirow{4}{*}{ETTm2} & 96 & 1.116 & 0.839 & \textbf{0.515} & \textbf{0.568} 
& 1.136 & 0.862 & \textbf{1.056} & \textbf{0.844} 
& 0.277 & 0.349 & \textbf{0.269} & \textbf{0.343} 
\\
 & 192 & 1.346 & 0.911 & \textbf{0.764} & \textbf{0.697} 
& 1.394 & 0.940 & \textbf{1.309} & \textbf{0.908} 
& 0.307 & \textbf{0.363} & \textbf{0.302} & 0.364 
\\
 & 336 & 1.573 & 1.040 & \textbf{1.117} & \textbf{0.858} 
& \textbf{1.640} & \textbf{1.055} & 1.661 & 1.086 
& 0.346 & 0.388 & \textbf{0.332} & \textbf{0.380} 
\\
 & 720 & \textbf{1.690} & \textbf{1.089} & 1.968 & 1.172 
& \textbf{2.164} & 1.248 & 2.265 & \textbf{1.243} 
& 0.421 & \textbf{0.424} & \textbf{0.414} & 0.428 
\\ \hline
\multirow{4}{*}{ETTh1} & 96 & 1.233 & 0.867 & \textbf{1.123} & \textbf{0.848} 
& 1.255 & 0.900 & \textbf{1.188} & \textbf{0.860} 
& \textbf{0.599} & \textbf{0.532} & 0.649 & 0.542 
\\
 & 192 & 1.307 & 0.913 & \textbf{1.172} & \textbf{0.858} 
& 1.276 & 0.907 & \textbf{1.157} & \textbf{0.822} 
& 0.675 & 0.560 & \textbf{0.582} & \textbf{0.519} 
\\
 & 336 & \textbf{1.253} & \textbf{0.892} & 1.270 & 0.912 
& 1.345 & 0.938 & \textbf{1.218} & \textbf{0.856} 
& 0.691 & 0.588 & \textbf{0.550} & \textbf{0.519} 
\\
 & 720 & 1.378 & \textbf{0.948} & 1.375 & 0.966 
& \textbf{1.154} & \textbf{0.843} & 1.191 & 0.857 
& 0.707 & 0.598 & \textbf{0.557} & \textbf{0.537} 
\\ \hline
\multirow{4}{*}{ETTh2} & 96 & \textbf{1.823} & \textbf{1.143} & 2.153 & 1.268 
& 2.224 & 1.273 & \textbf{1.882} & \textbf{1.099} 
& 0.378 & 0.425 & \textbf{0.373} & \textbf{0.421} 
\\
 & 192 & \textbf{2.128} & \textbf{1.238} & 2.745 & 1.444 
& \textbf{2.027} & \textbf{1.214} & 2.537 & 1.323 
& \textbf{0.404} & \textbf{0.439} & 0.411 & 0.451 
\\
 & 336 & 4.942 & 1.945 & \textbf{2.620} & \textbf{1.408} 
& 3.121 & 1.524 & \textbf{2.804} & \textbf{1.438} 
& \textbf{0.364} & \textbf{0.421} & 0.385 & 0.442 
\\
 & 720 & 3.508 & 1.627 & \textbf{2.661} & \textbf{1.345} & 3.924 & 1.712 & \textbf{3.054} & \textbf{1.492} & \textbf{0.435} & \textbf{0.467} & 0.449 & 0.478 \\
\hline
\end{tabular}%
}
\end{table*}

\begin{figure}[h]
    \centering
    \includegraphics[width=1.1\linewidth]{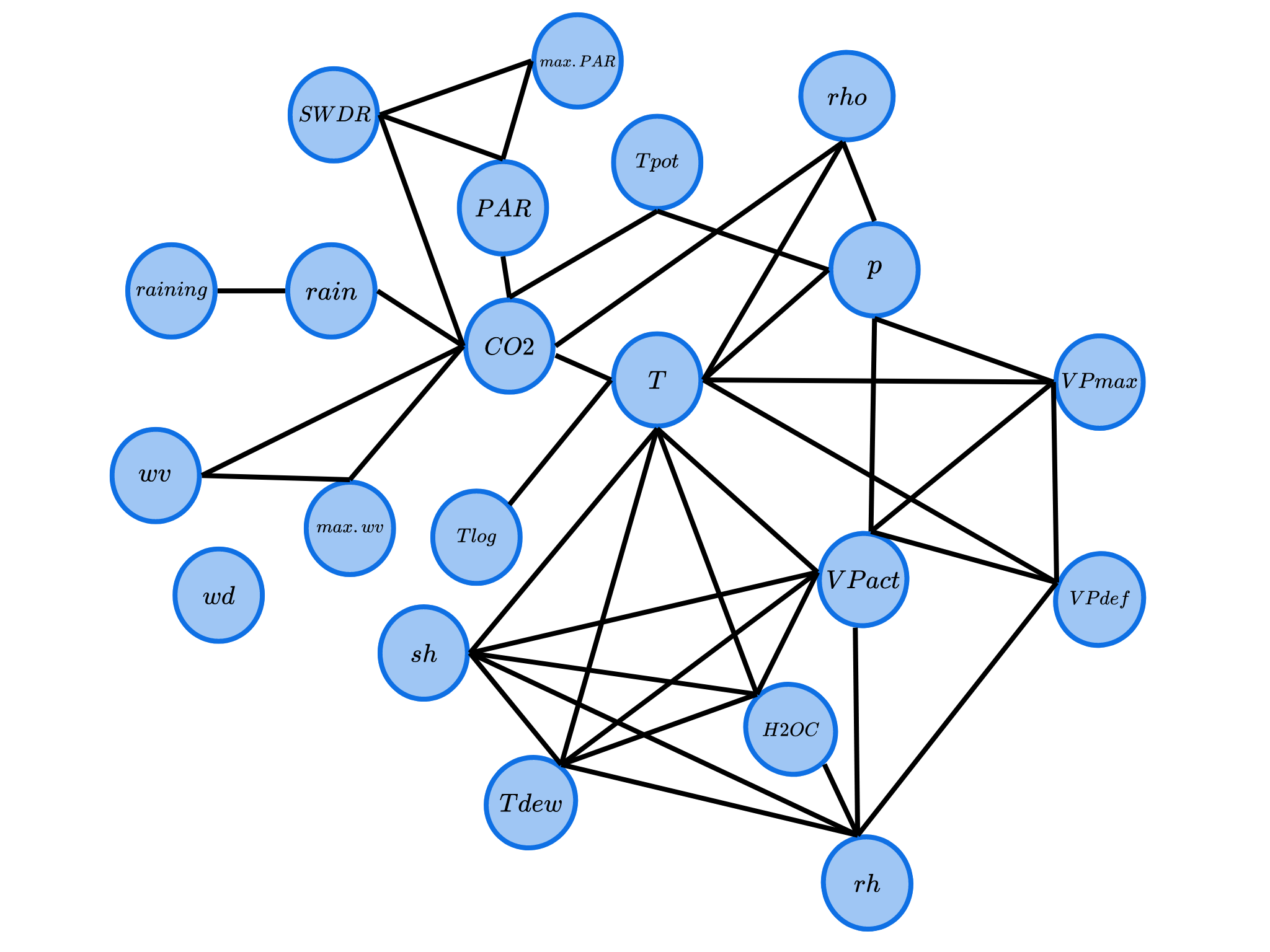}
    \caption{Knowledge Graph for Weather dataset; variable relationships are derived from theoretical laws and established empirical studies}
    \label{fig:4a}
\end{figure}

The construction of these embeddings incorporates insights from various sources: physical laws and observed empirical relationships for Weather \cite{weather_data} and operational principles for ETT \cite{informer}. These underlying principles are elaborated in the \textit{Appendix}, where we describe the specific relational frameworks used. For illustrative purposes, the intuitive knowledge graph for the Weather dataset is presented in Fig. \ref{fig:4a}. This structured presentation clearly outlines the foundational knowledge graphs integral to developing KGE.

\subsection{Experimental Setup}

We experiment with four popular architectures: Autoformer, Informer, PatchTST and Vanilla Transformer. Each of these architectures is distinguished by unique features, such as the decomposition blocks in Autoformer, ProbSparse attention by Informer, channel-independent patching by PatchTST and the canonical multi-head attention mechanism in Transformer. 
We replicate benchmark results for all the architectures and then incorporate our proposed KGE integration. For a precise comparison, we train all the models with the same experimental setup, viz. with a lookback window  (\(L\)) of 336 and across four horizons $H \in \{96, 192, 336, 720\}$. We maintain the consistency in configurations for finetuning of the original replication and the proposed ones. More details on the baseline configuration setup of each model are provided in the \textit{Appendix}. We evaluate these architectures on Mean Square Error (MSE) and Mean Absolute Error (MAE). 

\subsection{Results and Comparison}

Table \ref{tab3.1} presents the findings of a comprehensive comparative analysis between original model architectures and their enhanced iterations across varying horizon lengths. Within this multivariate framework, our methodology consistently surpasses benchmark results, particularly evident in the Weather dataset. Quantitatively, our proposed integration of knowledge graph embedding with Informer yields notable improvements, with a remarkable 54.5\% average reduction in Mean Squared Error (MSE) and a 39.3\% decrease in Mean Absolute Error (MAE). Similarly encouraging results are observed with  Transformer (22.6\% in MSE; 19.3\% in MAE) and Autoformer (7.9\% in MSE; 8\% in MAE). For PatchTST, we observe a 2.5\% reduction in MSE and 1.2\% in MAE for the Weather dataset as illustrated in Table \ref{tab3.2}.

\begin{table}[!]
\centering
\caption{Performance Comparison of PatchTST with and without Knowledge Graph Embeddings on Weather Dataset}
\label{tab3.2}
\begin{tabular}{|c||c|c|c|c||}
\hline
 \multirow{1}{*}{Horizon}     & \multicolumn{2}{c|}{\shortstack[c]{\\[0.1ex]PatchTST\\(With KGE)\\[0.1ex]}} & \multicolumn{2}{c||}{\shortstack[c]{\\[0.1ex]PatchTST\\(Original)\\[0.1ex]}
}\\                                        \cline{2-5}& \multicolumn{1}{l|}{MSE} & \multicolumn{1}{l|}{MAE} & \multicolumn{1}{l|}{MSE} & \multicolumn{1}{l||}{MAE
}\\ \hline

                           96& \textbf{0.148} & \textbf{0.199} & 0.152 & 0.200 
\\
  192 & \textbf{0.191} & \textbf{0.239} & 0.196 & 0.243 
\\
  336 & \textbf{0.241} & \textbf{0.280} & 0.248 & 0.283 
\\
  720 & \textbf{0.313} & \textbf{0.332} & 0.320 & 0.335 
\\ \hline
\end{tabular}%
\end{table}

Upon initial examination of the ETT dataset, striking enhancements stand out, particularly with ETTm1 across various models with notable improvements, exemplified by average reductions in MSE and MAE. For instance, Autoformer demonstrates a remarkable 21.9\% decrease in MSE and an 11.4\% decrease in MAE, Transformer shows average reductions of 2.5\% in MSE and 5.6\% in MAE, while Informer displays an improvement of 1\% in MSE and 1.2\% in MAE. Moreover, when focusing on ETTh2, the Autoformer model exhibits an average enhancement of 2.2\% in both MSE and MAE.

It is noteworthy that these performance boosts are more conspicuous when forecasting over longer horizons, highlighting the robustness of our approach necessary for handling extended prediction tasks within the ETT dataset. Conversely, sub-optimal results emerge when considering the impact of knowledge graph embeddings, particularly evident in ETTm2 and smaller datasets like ETTh1 and ETTh2. Here, we observe a muted influence; for instance, We observe a negligible improvement in the integration of KGE with PatchTST for ETT datasets. This could be potentially attributable to the simplicity of the knowledge graph structure in ETT and thus could lead to a redundant increment in model parameters by the introduction of KGEs.

\section{Conclusion}
\label{sec:conclusion}
In this study, we propose an intuitive knowledge graph embeddings (KGE) approach for enhancing the state-of-the-art (SOTA) architectures for the multivariate Long Sequence Timeseries Forecasting (LSTF) problem. These KGEs are learnable and dynamic as they are updated during the training of transformer-based architectures. These embeddings are derived from the adjacency matrix of the conceptual knowledge graph after they undergo a series of matrix transformations. Furthermore, they are fused with positional embeddings and other embeddings to train transformer-based architectures effectively. 

While purely dynamic graphs and their applications in timeseries forecasting have been extensively explored, integration of learnable knowledge graph embeddings in transformers through initial knowledge graphs remains underutilized. This paper demonstrates that a knowledge graph curated solely from conceptual relationships among variables can generate dynamic knowledge graph embeddings from time series data, delivering competitive results compared to benchmark models. Given the suboptimal performance of dynamic graph approaches for spatio-temporal LSTF compared to SOTA transformer variants. The initial basis for a knowledge graph is meticulously derived through a comprehensive analysis of potential conceptual relations among variables within the domain. We validate the performance of this integration through experimentation across various datasets with varying prediction lengths. After a comprehensive evaluation, it is evident that the dynamic KGE enhancement proved beneficial in understanding the inherent interdependencies among variables for LSTF.

While KGEs have demonstrated success, particularly on extensive datasets characterized by intuitive variable relationships, our proposed approach does present certain limitations that necessitate attention. The construction of a knowledge graph is inherently subjective and domain-specific. Thus, it is pertinent to recognize that this approach necessitates a thorough analysis of variables to establish relations. In such cases, KGE based on purely dynamic graph approaches may offer insights into variable relationships and could be a possible scope of research. However, it is crucial to emphasize that the primary objective of this study is to explore the potential of well-defined knowledge graphs for conceptual relations in variables that form the initial basis for dynamic KGEs. Nevertheless, for smaller datasets marked by significant temporal variations, the integration of KGE proves less advantageous. This study highlights that the introduction of KGEs has a clear scope of application wherever established relationships among variables exist.



\ifCLASSOPTIONcaptionsoff
  \newpage
\fi

\bibliographystyle{IEEEtran}
\bibliography{main}

\begin{thebibliography}{10}
\providecommand{\url}[1]{#1}
\csname url@samestyle\endcsname
\providecommand{\newblock}{\relax}
\providecommand{\bibinfo}[2]{#2}
\providecommand{\BIBentrySTDinterwordspacing}{\spaceskip=0pt\relax}
\providecommand{\BIBentryALTinterwordstretchfactor}{4}
\providecommand{\BIBentryALTinterwordspacing}{\spaceskip=\fontdimen2\font plus
\BIBentryALTinterwordstretchfactor\fontdimen3\font minus \fontdimen4\font\relax}
\providecommand{\BIBforeignlanguage}[2]{{%
\expandafter\ifx\csname l@#1\endcsname\relax
\typeout{** WARNING: IEEEtran.bst: No hyphenation pattern has been}%
\typeout{** loaded for the language `#1'. Using the pattern for}%
\typeout{** the default language instead.}%
\else
\language=\csname l@#1\endcsname
\fi
#2}}
\providecommand{\BIBdecl}{\relax}
\BIBdecl

\bibitem{mlp_forecasting}
P.~Zhang, Y.~Jia, J.~Gao, W.~Song, and H.~Leung, ``Short-term rainfall forecasting using multi-layer perceptron,'' \emph{IEEE Transactions on Big Data}, vol.~6, no.~1, pp. 93--106, 2020.

\bibitem{tsf_review}
B.~Lim and S.~Zohren, ``Time-series forecasting with deep learning: a survey,'' \emph{Philosophical Transactions of the Royal Society A}, vol. 379, no. 2194, p. 20200209, 2021.

\bibitem{tsf_dl_review}
Z.~Han, J.~Zhao, H.~Leung, K.~F. Ma, and W.~Wang, ``A review of deep learning models for time series prediction,'' \emph{IEEE Sensors Journal}, vol.~21, no.~6, pp. 7833--7848, 2021.

\bibitem{weather_research2}
S.~Kothapalli and S.~Totad, ``A real-time weather forecasting and analysis,'' in \emph{2017 IEEE International Conference on Power, Control, Signals and Instrumentation Engineering (ICPCSI)}.\hskip 1em plus 0.5em minus 0.4em\relax IEEE, 2017, pp. 1567--1570.

\bibitem{smart_grid}
K.~Wang, C.~Xu, Y.~Zhang, S.~Guo, and A.~Y. Zomaya, ``Robust big data analytics for electricity price forecasting in the smart grid,'' \emph{IEEE Transactions on Big Data}, vol.~5, no.~1, pp. 34--45, 2019.

\bibitem{energyload2}
G.~Hafeez, K.~S. Alimgeer, and I.~Khan, ``Electric load forecasting based on deep learning and optimized by heuristic algorithm in smart grid,'' \emph{Applied Energy}, vol. 269, p. 114915, 2020.

\bibitem{trafficflow1}
Y.~Hou, P.~Edara, and C.~Sun, ``Traffic flow forecasting for urban work zones,'' \emph{IEEE Transactions on Intelligent Transportation Systems}, vol.~16, no.~4, pp. 1761--1770, 2015.

\bibitem{trafficflow2}
I.~Lana, J.~Del~Ser, M.~Velez, and E.~I. Vlahogianni, ``Road traffic forecasting: Recent advances and new challenges,'' \emph{IEEE Intelligent Transportation Systems Magazine}, vol.~10, no.~2, pp. 93--109, 2018.

\bibitem{trafficflow3}
R.~Li, Y.~Hu, and Q.~Liang, ``T2f-lstm method for long-term traffic volume prediction,'' \emph{IEEE Transactions on Fuzzy Systems}, vol.~28, no.~12, pp. 3256--3264, 2020.

\bibitem{retail1}
S.~Punia, K.~Nikolopoulos, S.~P. Singh, J.~K. Madaan, and K.~Litsiou, ``Deep learning with long short-term memory networks and random forests for demand forecasting in multi-channel retail,'' \emph{International journal of production research}, vol.~58, no.~16, pp. 4964--4979, 2020.

\bibitem{financial1}
L.-J. Cao and F.~E.~H. Tay, ``Support vector machine with adaptive parameters in financial time series forecasting,'' \emph{IEEE Transactions on neural networks}, vol.~14, no.~6, pp. 1506--1518, 2003.

\bibitem{financial2}
M.~Wen, P.~Li, L.~Zhang, and Y.~Chen, ``Stock market trend prediction using high-order information of time series,'' \emph{Ieee Access}, vol.~7, pp. 28\,299--28\,308, 2019.

\bibitem{var}
J.~H. Stock and M.~W. Watson, ``Vector autoregressions,'' \emph{Journal of Economic perspectives}, vol.~15, no.~4, pp. 101--115, 2001.

\bibitem{gp}
W.~Yan, H.~Qiu, and Y.~Xue, ``Gaussian process for long-term time-series forecasting,'' in \emph{2009 international joint conference on neural networks}.\hskip 1em plus 0.5em minus 0.4em\relax IEEE, 2009, pp. 3420--3427.

\bibitem{svr}
D.~Basak, S.~Pal, D.~C. Patranabis \emph{et~al.}, ``Support vector regression,'' \emph{Neural Information Processing-Letters and Reviews}, vol.~11, no.~10, pp. 203--224, 2007.

\bibitem{nvidia}
T.~NVIDIA, ``Nvidia tesla v100 gpu architecture,'' \emph{Santa Clara, CA, USA}, 2017.

\bibitem{transformer}
A.~Vaswani, N.~Shazeer, N.~Parmar, J.~Uszkoreit, L.~Jones, A.~N. Gomez, {\L}.~Kaiser, and I.~Polosukhin, ``Attention is all you need,'' \emph{Advances in neural information processing systems}, vol.~30, 2017.

\bibitem{informer}
H.~Zhou, S.~Zhang, J.~Peng, S.~Zhang, J.~Li, H.~Xiong, and W.~Zhang, ``Informer: Beyond efficient transformer for long sequence time-series forecasting,'' in \emph{Proceedings of the AAAI conference on artificial intelligence}, vol.~35, no.~12, 2021, pp. 11\,106--11\,115.

\bibitem{autoformer}
H.~Wu, J.~Xu, J.~Wang, and M.~Long, ``Autoformer: Decomposition transformers with auto-correlation for long-term series forecasting,'' \emph{Advances in neural information processing systems}, vol.~34, pp. 22\,419--22\,430, 2021.

\bibitem{fedformer}
T.~Zhou, Z.~Ma, Q.~Wen, X.~Wang, L.~Sun, and R.~Jin, ``Fedformer: Frequency enhanced decomposed transformer for long-term series forecasting,'' in \emph{International conference on machine learning}.\hskip 1em plus 0.5em minus 0.4em\relax PMLR, 2022, pp. 27\,268--27\,286.

\bibitem{patchtst}
Y.~Nie, N.~H. Nguyen, P.~Sinthong, and J.~Kalagnanam, ``A time series is worth 64 words: Long-term forecasting with transformers,'' \emph{arXiv preprint arXiv:2211.14730}, 2022.

\bibitem{weather_data}
``Weather dataset: Max planck institute for biogeochemistry, germany,'' \url{https://www.bgc-jena.mpg.de/wetter/}, accessed: 2024-02-10.

\bibitem{traffic_data}
``Pems data source: California department of transportation,'' \url{https://pems.dot.ca.gov/}, accessed: 2024-02-10.

\bibitem{electricity_data}
A.~Trindade, ``{ElectricityLoadDiagrams20112014},'' UCI Machine Learning Repository, 2015, {DOI}: https://doi.org/10.24432/C58C86.

\bibitem{ILI_data}
``Ili dataset: National, regional, and state level outpatient illness and viral surveillance,'' \url{https://gis.cdc.gov/grasp/fluview/fluportaldashboard.html}, accessed: 2024-02-10.

\bibitem{att_failure_tsf}
B.~Lim and S.~Zohren, ``Time-series forecasting with deep learning: a survey,'' \emph{Philosophical Transactions of the Royal Society A}, vol. 379, no. 2194, p. 20200209, 2021.

\bibitem{relphormer}
Z.~Bi, S.~Cheng, J.~Chen, X.~Liang, F.~Xiong, and N.~Zhang, ``Relphormer: Relational graph transformer for knowledge graph representations,'' \emph{Neurocomputing}, vol. 566, p. 127044, 2024.

\bibitem{spacetimetransformer}
J.~Grigsby, Z.~Wang, N.~Nguyen, and Y.~Qi, ``Long-range transformers for dynamic spatiotemporal forecasting,'' \emph{arXiv preprint arXiv:2109.12218}, 2021.

\bibitem{DKG_mLSTF}
X.~Wang, Y.~Wang, J.~Peng, and Z.~Zhang, ``Multivariate long sequence time-series forecasting using dynamic graph learning,'' \emph{Journal of Ambient Intelligence and Humanized Computing}, vol.~14, no.~6, pp. 7679--7693, 2023.

\bibitem{DKG_mLSTF1}
W.~Shao, Z.~Jin, S.~Wang, Y.~Kang, X.~Xiao, H.~Menouar, Z.~Zhang, J.~Zhang, and F.~Salim, ``Long-term spatio-temporal forecasting via dynamic multiple-graph attention,'' \emph{arXiv preprint arXiv:2204.11008}, 2022.

\bibitem{kde_gan_lstf}
Y.~Fang, Y.~Qin, H.~Luo, F.~Zhao, and K.~Zheng, ``Stwave+: A multi-scale efficient spectral graph attention network with long-term trends for disentangled traffic flow forecasting,'' \emph{IEEE Transactions on Knowledge and Data Engineering}, 2023.

\bibitem{kde_dynamic_graph}
M.~Jin, Y.~Zheng, Y.-F. Li, S.~Chen, B.~Yang, and S.~Pan, ``Multivariate time series forecasting with dynamic graph neural odes,'' \emph{IEEE Transactions on Knowledge and Data Engineering}, 2022.

\bibitem{kde_gnn}
L.~Chen, D.~Chen, Z.~Shang, B.~Wu, C.~Zheng, B.~Wen, and W.~Zhang, ``Multi-scale adaptive graph neural network for multivariate time series forecasting,'' \emph{IEEE Transactions on Knowledge and Data Engineering}, 2023.

\bibitem{boxj}
G.~E. Box, G.~M. Jenkins, G.~C. Reinsel, and G.~M. Ljung, \emph{Time series analysis: forecasting and control}.\hskip 1em plus 0.5em minus 0.4em\relax John Wiley \& Sons, 2015.

\bibitem{expsmooth}
R.~G. Brown and R.~F. Meyer, ``The fundamental theorem of exponential smoothing,'' \emph{Operations Research}, vol.~9, no.~5, pp. 673--685, 1961.

\bibitem{STL}
C.~RB, ``Stl: A seasonal-trend decomposition procedure based on loess,'' \emph{J Off Stat}, vol.~6, pp. 3--73, 1990.

\bibitem{BSTS}
S.~L. Scott and H.~R. Varian, ``Predicting the present with bayesian structural time series,'' \emph{International Journal of Mathematical Modelling and Numerical Optimisation}, vol.~5, no. 1-2, pp. 4--23, 2014.

\bibitem{DLM}
M.~West and J.~Harrison, \emph{Bayesian forecasting and dynamic models}.\hskip 1em plus 0.5em minus 0.4em\relax Springer Science \& Business Media, 2006.

\bibitem{pyraformer}
S.~Liu, H.~Yu, C.~Liao, J.~Li, W.~Lin, A.~X. Liu, and S.~Dustdar, ``Pyraformer: Low-complexity pyramidal attention for long-range time series modeling and forecasting,'' in \emph{International conference on learning representations}, 2021.

\bibitem{LSTNET}
G.~Lai, W.-C. Chang, Y.~Yang, and H.~Liu, ``Modeling long-and short-term temporal patterns with deep neural networks,'' in \emph{The 41st international ACM SIGIR conference on research \& development in information retrieval}, 2018, pp. 95--104.

\bibitem{reformer}
N.~Kitaev, {\L}.~Kaiser, and A.~Levskaya, ``Reformer: The efficient transformer,'' \emph{arXiv preprint arXiv:2001.04451}, 2020.

\bibitem{attentionfree}
H.~Inzirillo and L.~De~Villelongue, ``An attention free long short-term memory for time series forecasting,'' \emph{arXiv preprint arXiv:2209.09548}, 2022.

\bibitem{improvedposencoding}
N.~M. Foumani, C.~W. Tan, G.~I. Webb, and M.~Salehi, ``Improving position encoding of transformers for multivariate time series classification,'' \emph{Data Mining and Knowledge Discovery}, vol.~38, no.~1, pp. 22--48, 2024.

\bibitem{cape}
T.~Likhomanenko, Q.~Xu, G.~Synnaeve, R.~Collobert, and A.~Rogozhnikov, ``Cape: Encoding relative positions with continuous augmented positional embeddings,'' \emph{Advances in Neural Information Processing Systems}, vol.~34, pp. 16\,079--16\,092, 2021.

\bibitem{roformer}
J.~Su, M.~Ahmed, Y.~Lu, S.~Pan, W.~Bo, and Y.~Liu, ``Roformer: Enhanced transformer with rotary position embedding,'' \emph{Neurocomputing}, vol. 568, p. 127063, 2024.

\bibitem{nonstationarytransformer}
Y.~Liu, H.~Wu, J.~Wang, and M.~Long, ``Non-stationary transformers: Exploring the stationarity in time series forecasting,'' \emph{Advances in Neural Information Processing Systems}, vol.~35, pp. 9881--9893, 2022.

\bibitem{GNN}
F.~Scarselli, M.~Gori, A.~C. Tsoi, M.~Hagenbuchner, and G.~Monfardini, ``The graph neural network model,'' \emph{IEEE transactions on neural networks}, vol.~20, no.~1, pp. 61--80, 2008.

\bibitem{GCN}
T.~N. Kipf and M.~Welling, ``Semi-supervised classification with graph convolutional networks,'' \emph{arXiv preprint arXiv:1609.02907}, 2016.

\bibitem{kde_gan}
R.~Li, F.~Zhang, T.~Li, N.~Zhang, and T.~Zhang, ``Dmgan: Dynamic multi-hop graph attention network for traffic forecasting,'' \emph{IEEE Transactions on Knowledge and Data Engineering}, 2022.

\bibitem{dyn_graph_nn}
V.~La~Gatta, V.~Moscato, M.~Postiglione, and G.~Sperlí, ``An epidemiological neural network exploiting dynamic graph structured data applied to the covid-19 outbreak,'' \emph{IEEE Transactions on Big Data}, vol.~7, no.~1, pp. 45--55, 2021.

\bibitem{dynamic_hypergraph}
S.~Wang, Y.~Zhang, X.~Lin, Y.~Hu, Q.~Huang, and B.~Yin, ``Dynamic hypergraph structure learning for multivariate time series forecasting,'' \emph{IEEE Transactions on Big Data}, vol.~10, no.~4, pp. 556--567, 2024.

\bibitem{DSTAGCN}
Q.~Zheng and Y.~Zhang, ``Dstagcn: Dynamic spatial-temporal adjacent graph convolutional network for traffic forecasting,'' \emph{IEEE Transactions on Big Data}, vol.~9, no.~1, pp. 241--253, 2023.

\bibitem{stgcn}
B.~Yu, H.~Yin, and Z.~Zhu, ``Spatio-temporal graph convolutional networks: A deep learning framework for traffic forecasting,'' \emph{arXiv preprint arXiv:1709.04875}, 2017.

\bibitem{stemgnn}
D.~Cao, Y.~Wang, J.~Duan, C.~Zhang, X.~Zhu, C.~Huang, Y.~Tong, B.~Xu, J.~Bai, J.~Tong \emph{et~al.}, ``Spectral temporal graph neural network for multivariate time-series forecasting,'' \emph{Advances in neural information processing systems}, vol.~33, pp. 17\,766--17\,778, 2020.

\bibitem{attention_gcn}
Y.~Zhang, X.~Wei, X.~Zhang, Y.~Hu, and B.~Yin, ``Self-attention graph convolution residual network for traffic data completion,'' \emph{IEEE Transactions on Big Data}, vol.~9, no.~2, pp. 528--541, 2023.

\bibitem{graphtransformer}
V.~P. Dwivedi and X.~Bresson, ``A generalization of transformer networks to graphs,'' \emph{arXiv preprint arXiv:2012.09699}, 2020.

\bibitem{graphformer}
Y.~Wang, H.~Long, L.~Zheng, and J.~Shang, ``Graphformer: Adaptive graph correlation transformer for multivariate long sequence time series forecasting,'' \emph{Knowledge-Based Systems}, vol. 285, p. 111321, 2024.

\bibitem{dms}
G.~Chevillon, ``Direct multi-step estimation and forecasting,'' \emph{Journal of Economic Surveys}, vol.~21, no.~4, pp. 746--785, 2007.

\bibitem{ims_vs_dms1}
M.~W. McCracken and J.~T. McGillicuddy, ``An empirical investigation of direct and iterated multistep conditional forecasts,'' \emph{Journal of Applied Econometrics}, vol.~34, no.~2, pp. 181--204, 2019.

\bibitem{dlinear}
A.~Zeng, M.~Chen, L.~Zhang, and Q.~Xu, ``Are transformers effective for time series forecasting?'' in \emph{Proceedings of the AAAI conference on artificial intelligence}, vol.~37, no.~9, 2023, pp. 11\,121--11\,128.

\end{thebibliography}

\end{document}